\documentclass[journal]{IEEEtran}
\usepackage{url}
\usepackage{subfig}
\usepackage{graphicx}
\usepackage{algorithm}
\usepackage{algpseudocode}
\usepackage{booktabs}
\usepackage{booktabs}
\usepackage{makecell}
\usepackage{amsmath}
\usepackage{amsfonts}
\usepackage{threeparttable}

\ifCLASSINFOpdf
\else
\fi
\begin{document}
%
\title{EdgeSync: Faster Edge-model Updating via Adaptive Continuous Learning for Video Data Drift}
%
%
%

\author{Peng Zhao,~\IEEEmembership{Member,~IEEE,}
        Runchu Dong, Guiqin Wang, 
        and~Cong Zhao,~\IEEEmembership{Member,~IEEE,}
\thanks{Peng Zhao, Runchu Dong, and Guiqing Wang are with School of Computer Science and Technology, and with the National Engineering Laboratory for Big Data Analytics (NEL-BDA), Xi’an Jiaotong University, China. e-mail: p.zhao@mail.xjtu.edu.cn.}
\thanks{Cong Zhao is with School of Mathematics and Statistics,and with the National Engineering Laboratory for Big Data Analytics (NEL-BDA), Xi’an Jiaotong University, China.}}

%
%

\markboth{Journal of \LaTeX\ Class Files,~Vol.~14, No.~8, August~2015}%
{Shell \MakeLowercase{\textit{et al.}}: Bare Demo of IEEEtran.cls for IEEE Journals}
%



\maketitle

\begin{abstract}
Real-time video analytics systems typically place models with fewer weights on edge devices to reduce latency. The distribution of video content features may change over time for various reasons (i.e. light and weather change) , leading to accuracy degradation of existing models, to solve this problem, recent work proposes a framework that uses a remote server to continually train and adapt the lightweight model at edge with the help of complex model. However, existing analytics approaches leave two challenges untouched: firstly, retraining task is compute-intensive, resulting in large model update delays; secondly, new model may not fit well enough with the data distribution of the current video stream. To address these challenges, in this paper, we present EdgeSync, EdgeSync filters the samples by considering both timeliness and inference results to make training samples more relevant to the current video content as well as reduce the update delay, to improve the quality of training, EdgeSync also designs a training management module that can efficiently adjusts the model training time and training order on the runtime. By evaluating real datasets with complex scenes, our method improves about 3.4\% compared to existing methods and about 10\% compared to traditional means.
\end{abstract}


%
\IEEEpeerreviewmaketitle

\section{Introduction}
%
%
%
%

\IEEEPARstart{R}{eal-time} video analytics has significant potential for applications in various fields, such as augmented reality, video surveillance, and traffic detection\cite{ananthanarayanan2017real}. Owing to recent advancements in deep neural networks (DNNs), the performance of video analysis has been greatly enhanced, even surpassing human accuracy in many scenarios\cite{he2016deep,chen2023diffusiondet,zhao2017pyramid}. Although advanced DNNs can generate accurate inferences for various challenging analytical tasks, the complexity of the network structures and the large number of parameters increase the computational burden\cite{dosovitskiy2020image}. This makes it challenging for these models to perform real-time analytics on resource-constrained devices, such as mobile terminals and edge devices.

To meet the demands of real-time analysis on edge devices, deep neural networks with fewer weights and shallower architectures are typically deployed. However, the distribution of video content in real scenarios often changes over time (e.g., variations in lighting, crowd density, and weather conditions), making lightweight models vulnerable to data drift\cite{smith2023closer}\cite{tan2019efficientnet}. Consequently, it is challenging to maintain the desired accuracy with a model obtained by single offline training. Continuous learning has recently been validated as a feasible solution to enhance the adaptability of edge models. For instance, Mullapudi \textit{et al}. in \cite{mullapudi2019online} demonstrated the effectiveness of this approach by proposing an online model distillation technique to train a low-cost student model on live video streaming. Ekya\cite{bhardwaj2022ekya} refined this framework by performing the data annotation in the cloud and jointly adjusting real-time inference and continuous learning based model retraining on edge servers to maximize the overall accuracy through a scheduler.

\begin{figure}[!t]
\centering
\subfloat[Scene Change Example]{\includegraphics[width=3.3in]{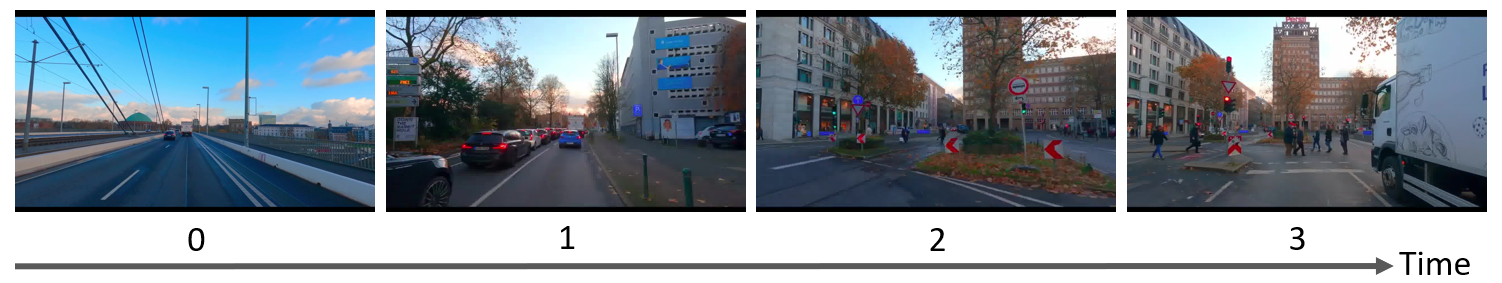}%
\label{fig_example_1}}
\hfil
\subfloat[Accuracy]{\includegraphics[width=1.65in]{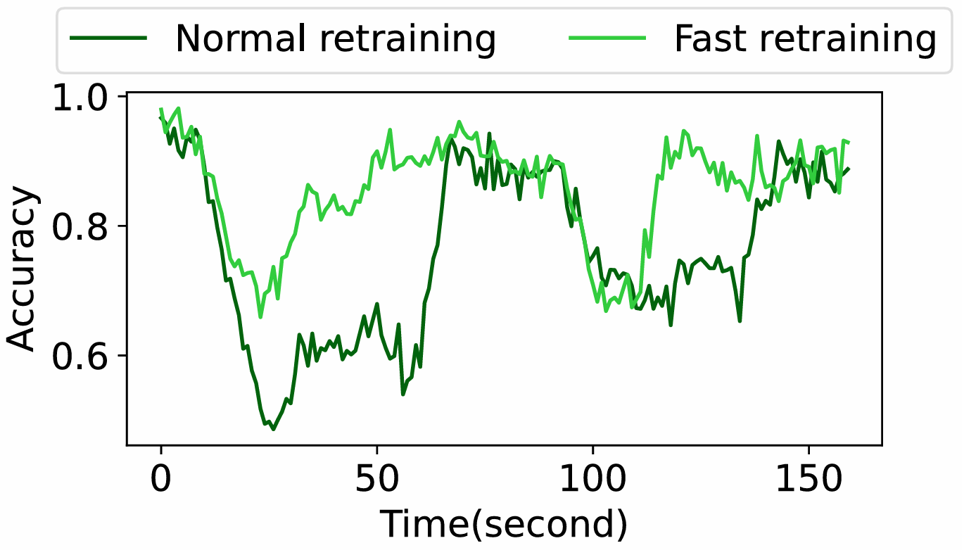}%
\label{fig_example_2}}
\subfloat[Class Distribution]{\includegraphics[width=1.65in]{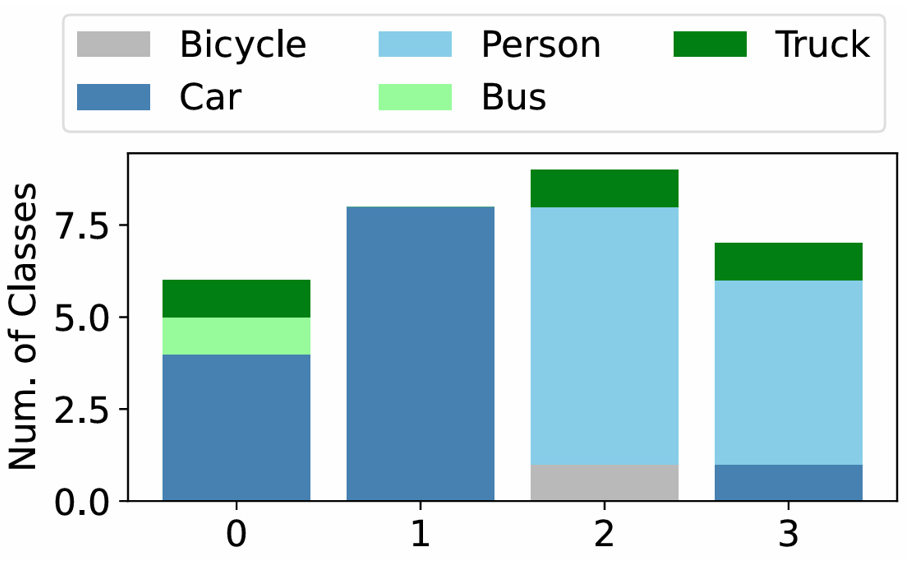}%
\label{fig_example_3}}
\caption{Changes of class distributions and accuracy when a car camera enters downtown of ideal fast retraining and normal retraining.}
\label{fig_example}
\end{figure}

Though several previous work have recently been conducted to enhance the accuracy of edge models through online continuous learning, they still suffer from the challenge of large update delay of updating new edge models, result in the decrease of overall inference accuracy. First, lightweight models at the edge continue to use outdated parameters for inference while retraining new ones in the cloud. Some methods either upload all samples to the cloud for labeling and retraining or perform additional operations such as hyperparameter selection in the cloud, which consumes considerable time and increases the update latency to data drift, as shown in Fig. \ref{fig_example}. Second, current methods mainly set a static sampling rate and a fixed time interval for model updating, since video streams vary in different degrees over time, they do not take into account the degree of influence of different samples and the upper limit of the model's own accuracy in the current time period, thus affecting the quality of the model's retraining and decreasing the generalization of the retrained model.

Although several previous work has maintained the accuracy of edge devices, several challenges remain. First, the update delay of new edge models is large. Lightweight models at the edge continue to use outdated parameters for inference while retraining new ones. Current methods either upload all samples to the cloud for labeling and retraining or perform additional operations such as hyperparameter selection in the cloud, which consumes considerable time and increases the response latency to data drift, as shown in Figure 1. Second, current methods primarily use a static sampling rate and a fixed time interval for model updates. Since video streams vary to different extents over time, these methods do not consider the varying impact of different samples or the model's upper accuracy limit in the current period. This oversight affects the quality of model retraining and decreases the generalization of the retrained model.


In order to solve these problems, in this paper, we propose EdgeSync, an Faster Edge-model Updating approach that automatically and continually adapts models based on the scene. It consists of two modules: a sample filtering module and a model retraining management module. Specifically, to reduce the network uploading bandwidth as well as improve the quality of training samples, the sample filtering module filters certain samples of the current video stream in real time, taking into account the model's ability to extract features from samples and the importance of timeliness. To further ensure the quality of model training while accelerating its update speed, we propose the model retraining management module, which efficiently makes accurate training handover decisions with low complexity. Benefiting from these two modules, EdgeSync can achieve fast model updates in real scenes, which helps improve the accuracy of single camera inference and allows the cloud to carry out more camera model update tasks.

The major contributions of this paper are summarized as follows:
\begin{itemize}
  \item First, we propose a fast method for filtering video streaming samples, which combines the results of timeliness and adaptability to filter out unnecessary samples. Benefiting from the module, We can flexibly adjust the number of samples according to network conditions while improving the quality of samples used for training in the cloud.
  \item Next, we propose a re-training manager to adjust the training order and training time by using the labeled and trained computed features, to further improve the model update speed, we use offline and online profiling to accelerate hyper-parameter selection procedure.
  \item Third, we conduct extensive experiments to evaluate the performance of EdgeSync in real-world scenarios. Evaluation results demonstrate that EdgeSync can reduce network bandwidth and speed up model update frequencies. It also outperforms the baseline schemes in terms of overall accuracy.
\end{itemize}

The structure of the paper is outlined as follows: In Section 2, we review related work. Section 3 introduces the system architecture and details the sample filtering module and retraining management module. Section 4 presents the evaluation results. Finally, concluding remarks are provided in Section 5.


\section{Related Work}

\subsection{Video Analytics Systems}
Real-time video analytics uses computer vision algorithms to automatically analyze and understand the content of video streams generated by one or more cameras, thus accomplishing complex tasks such as target recognition and anomaly detection while the video streams are being recorded and transmitted. To balance accuracy and speed, Chameleon\cite{jiang2018chameleon} designs a controller to dynamically select parameters in the video analysis system; Wang \MakeLowercase{\textit{et al.}}\cite{wang2023edge} addresses the joint configuration tuning problem for video parameters and serverless computation resources, they propose an algorithm utilizing Markov approximation to select optimal configurations for video streams by considering the accuracy-cost trade-off. PacketGame\cite{yuan2023packetgame} selectively filters packets through a specially designed neural network before running the video decoder in order to increase the number of videos to be processed in parallel; Ekya\cite{bhardwaj2022ekya} performs simultaneous training and inference at the edge server, and designs a thief scheduler to select the appropriate parameters to balance training and inference. AMS\cite{khani2021real} and JIT\cite{mullapudi2019online} create specialized lightweight DNNs to maintain accuracy under a specific scene, the challenge is that as the video scene changes, the system must dynamically create new DNNs to adapt to the new video content, our work is similar with them and focuses on an area that has not been emphasized: the update quality and speed of specialized lightweight DNNs, we hope to accelerate the model's ability to adapt to the video and improve the overall accuracy by improving the efficiency of model updating.

\subsection{Continuous Learning}

Continuous learning (also called incremental learning) utilizes new data acquired to continuously adapt the current model to new tasks while remember basic concepts previously learned\cite{de2021continual}. Recent works have addressed catastrophic forgetting with longer task sequences, \cite{wang2022learning} employ a meta optimizer that dynamically adapts the learning rate to prevent forgetting throughout the learning process in SDML. GPM\cite{saha2020gradient} identifies crucial gradient subspaces related to previous tasks and mitigates catastrophic forgetting by implementing gradient steps orthogonal to these subspaces when acquiring knowledge in a new task. Some of them focus on limited training samples, MAML\cite{finn2017model} tunes the model's parameters through training via gradient descent on a new task. DeepBDC\cite{xiao2022few} acquires image representations by assessing the disparity between joint characteristic functions of embedded features and the product of their marginals.

Existing efforts toward learning new tasks continually without forgetting the past tasks mainly classified into three categories. The first group  dedicate different subsets of network parameters to each task\cite{rusu2016progressive}\cite{yan2021dynamically}, the second group attempt to overcome forgetting in fixed capacity model through structural regularization which penalizes major changes in the parameters that were
important for the previous tasks\cite{kirkpatrick2017overcoming}\cite{hou2019learning}. The third group of method mitigate forgetting by either storing a subset
of (raw) examples from the past tasks in the memory for rehearsal\cite{lopez2017gradient}\cite{bang2021rainbow} or synthesizing old data from generative models to perform pseudo rehearsal\cite{shin2017continual}. Our method shares the idea with continuous learning, but uses a different approach and framework, we let lightweight DNNs focus more on recent video frames based on spatial locality relation and train it on the fly.

\subsection{Unsupervised Adaptation Methods}

\begin{figure*}[t]
\centering
\includegraphics[width=6.21in]{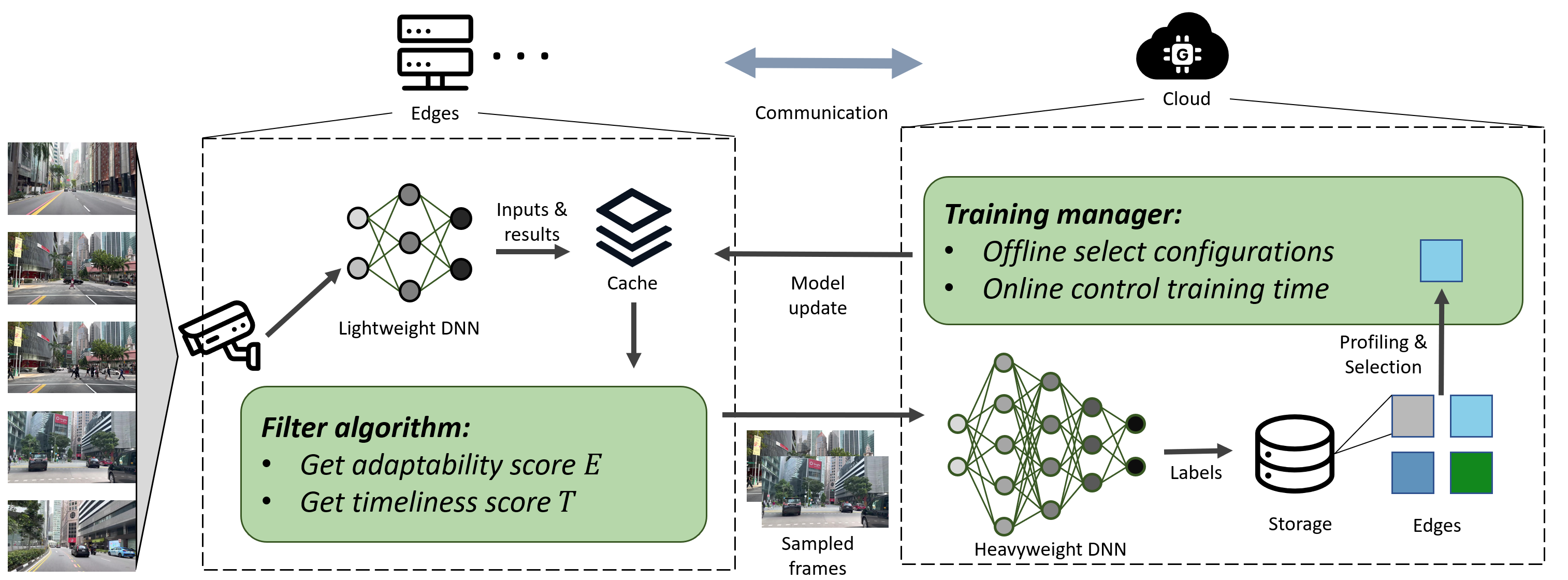}
\caption{Overall architecture of EdgeSync}
\label{fig_overview}
\end{figure*}

Unsupervised adaptation methods aim to improve accuracy to tackle potential distribution shifts between training and testing data. Early unsupervised domain adaptation methods fine-tune the model rely on both source domain and target domain\cite{long2016unsupervised}\cite{kang2019contrastive}. Some work attempts to solve this problem based only on target domain. Sahoo \MakeLowercase{\textit{et al.}}\cite{chidlovskii2016domain} propose a method which only requires unlabeled test data, SHOT\cite{liang2020we}  combined entropy minimization with pseudo labeling, TTT\cite{sun2020test} turns a single unlabeled test sample into a self-supervised learning problem, then update the model parameters before making a prediction. To reduce the training time, Tent\cite{chen2022contrastive} optimize the model for confidence as measured by test entropy minimization, LAME\cite{Boudiaf_2022_CVPR} adapts the model's output rather than its parameters by discouraging deviations from the prediction to find the optimal set of latent assignments which mitigating the effect of hyperparameters on performance. 

In addition, a few work has been proposed tailored for video data. Zeng \MakeLowercase{\textit{et al.}}\cite{zeng2023exploring} propose a new test-time learning scheme that leverages motion cues in videos to enhance the generalization capability of video classification models. Lin\cite{lin2023video} propose a test-time adaptation method for video action recognition models, it aligns the training statistics with the online estimates of target statistics, then enforce prediction consistency among temporally augmented views of a video sample. Our approach leverage a complex model in the cloud server to do supervised training with its generated labels, which is which is more suitable in this framework.

\section{Method}

In this section, we present the proposed method EdgeSync for real-time video analytics. We first introduce the overall system architecture of EdgeSync. Following that, we present the details of our approach, including the sample filtering module and the adaptive model training module.

\subsection{System Architecture}

Fig.\ref{fig_overview} illustrates the overall architecture of EdgeSync, an edge-cloud collaborative system for real-time video analytics. Basically, this system consists of multiple terminal cameras, several edge servers, and a centralized cloud server. Each edge device is primarily responsible for locally analyzing video streams from one or more cameras utilizing a lightweight model. In contrast, the cloud server is responsible for model management across all edges, including model training and updates. Specifically, the fundamental modules within both the edge and cloud servers can be summarized as follows.

\textbf{Edge server}: Each edge deploys a lightweight model with a shallow architecture and few parameters to perform the rapid local inference of real-time video streaming. To maintain the inference accuracy and address the challenge of data drift, the inference results and corresponding video frames are then buffered in a local cache for online continuous learning. Since sending all frames directly to the cloud will introduce significant network overhead, the edge side must selectively upload samples. In this study, we proposed a sample filtering module, which is used to score the quality of the samples in the current period of time and dynamically select high-quality samples for training the new model at the cloud. In particular, this module scores the quality of the current samples from two aspects. First, it uses the output confidence distribution derived from the current model deployed at the edge, calculating the entropy of the distribution. Higher entropy produces more update information for the current model. Then, it takes into account the temporal distribution. The distance from the current time point represents the similarity with the current sample period. The edge then sorts these samples based on the above two aspects and finally selects the top several samples to transfer their results and frames to the cloud for retraining. On the other hand, each edge also receives new model from the cloud server for updating. 

\textbf{Cloud server}: The cloud utilizes frames sent from the edges to dynamically train new edge models and then dispatches the updates to the edges. Upon receiving the data samples from the edge, a training management module is designed in the cloud to perform two sequential processes: labeling and retraining. In particular, to label new data samples, the cloud employs a complex heavyweight model, leveraging its highly accuracy and abundant resources.This complex model generates highly accurate predicted values, which are treated as ground truth labels to supervise the training of the lightweight edge models. The labeled data are then stored in a storage buffer. Before retraining, the cloud evaluates the historical accuracy of each edge to prioritize which one to train. During the training phase, the cloud uses pre-selected hyperparameters and employs an early-stopping mechanism to accelerate the training process. Finally, the updated model is sent back to the corresponding edge. Notably, only the updated parameters, rather than the entire edge model, are replaced. Additionally, the model and its parameters are stored in GPU memory to further accelerate model updating and reduce delays in context switching between the GPU and memory.

\subsection{Sample Filtering Module}

In practice, it is unnecessary and time-consuming for the edge to process the inputs simultaneously. This is because video frames can provide abundant information, yet not all necessarily contribute to improving model retraining performance. To improve the quality and accelerate the sending procedure, we propose a sample filtering module to selectively upload samples at edge.

In the following, we present the detailed definitions. There are $k$ edges, each of them contains a video stream. For each video stream $v \in V$ in an update window T, we decide: (1) the model $f$ deployed in each edge, whose parameter is $\theta$;(2) the frame $x_i$ in each window($i$ represents sequence in current window); (2) the inference result $y_i$ for frame $x_i$ using current model $f$; and (4) cache $Y$ to store inference result and related frame, cache $M$ to store frame quality. It’s worth noting that the size of update window $T$ changes every time.

In the filtering module, we address two key aspects. Firstly, we evaluate the adaptability of the edge device model to the current sample. If the current model has accurately predicted the sample, retraining is unnecessary. Conversely, samples that the current model performs poorly on necessitate retraining. Secondly, we consider the timeliness of the current sample. Given our focus on enhancing lightweight model accuracy through adaptation to local frames, it is vital to select samples that reflect the current video stream distribution. The filter must prioritize samples closer to the current timestamp while ensuring an adequate amount of retraining data.

For the adaptability score $E\left(x;\theta\right)$, we do not use the model output confidence because modern neural networks are poorly calibrated \cite{DBLP:conf/icml/GuoPSW17}, instead we use the entropy of the model outputs as a basis for adaptability judgment $E$, as it has achieved good results in other fields:
\begin{equation} \label{eq:1}
E\left(x;\theta\right)=-f_\theta(y|x) \cdot \log{f_\theta(y|x)}
\end{equation}
Where x denotes the current sample, y denotes the result predicted by model f in current parameter $\theta$. The higher the adaptability score, The lower the model's certainty for the current sample. For the timeliness score $I(x)$, we use the following formula to represent the timeliness importance:
\begin{equation} \label{eq:2}
I\left(i\right)=1/(1+\exp(-i/T))
\end{equation}
where $T$ denotes the size of current window, and $i$ denotes the relative index of the arrival of sample $x$ to current timestamp. The index of current sample is initialized as 0, so older samples tend to have bigger index $i$.

Finally we use weighted average to sum up two scores:
\begin{equation} \label{eq:3}
Q\left(x, i, \alpha, \beta\right)=\alpha E\left(x;\theta\right)+\beta I\left(i\right)
\end{equation}
In the formula, $\alpha$ and $\beta$ are used to balance adaptability and timeliness.

\begin{algorithm}
\caption{Sample Filtering Algorithm}\label{alg:cap}
\begin{algorithmic}[1]
\State \textbf{Input} samples $Y \gets \{(x_1, y_1), (x_2, y_2), ..., (x_n, y_n)\}$ in cache 
\State \textbf{Define} update window size $T$, Percentage of uploaded samples $k$, $\alpha$, $\beta$ \newline
list $M \gets, \emptyset$
\For {$(i, x, y) \in Y$}
    \State $adaptability_{i} \gets E(x;\Theta)$, \Comment{Calculate adaptability score via \ref{eq:1}}
    \State $timeliness_{i} \gets I(i)$ \Comment{Calculate timeliness score via \ref{eq:2}}
    \State $quality_{i} \gets Q(x, i, \alpha, \beta)$ \Comment{Get overall score for $x$ via \ref{eq:3}}
    \State append $(quality_{i}, x, y, i)$ to list $M$
\EndFor
\State $M \gets sort(M)$ sort element in $M$ by $quality_{i}$
\State $M \gets$ get top-k percentage elements from sorted list $M$ and remove element $quality$
\State send $M$ to cloud
\end{algorithmic}
\end{algorithm}

The pseudo code in Algorithm 1 summarizes the process of sample filtering module. Subsequently, the inference results \( y \) and corresponding video frames \( x \) are buffered in a local cache to facilitate online continuous learning. The module assesses the quality of current samples from two perspectives. Firstly, it computes the adaptability score using Formula 1, followed by the calculation of the timeliness score using Formula 2. Based on these criteria, the edge device ranks the samples and selects the top few for transmitting their results and frames to the cloud for retraining.

\subsection{Training Manager}
The lightweight model at the edge maintains its accuracy for its video stream through continuous training, we dynamically update the model parameters and adjust training sequence through the retraining managing module to maximize the overall accuracy.

\subsubsection{Select a model for retraining}
Different edge may have different degrees of change in video content, for example, When monitoring road conditions, engineers will place cameras in different locations in the city for monitoring. Due to different geographical locations, the video streams faced by each camera have different change patterns, Therefore, we need to make profiling on these video streams and select a appropriate one for retraining in order to let the edge adapt this current local distribution faster which scene change more difficultly and frequently.

The cloud receives some samples and corresponding inference results transmitted from the edge. We propose to build up profiling problem based on the model inference result level rather than the sample feature level, as it provides a robuster signal for measuring degrees of scene changes\cite{khani2021real}, as well as avoiding additional computing overhead like some methods, which add a feature extractor at cloud. We then choose to use history window-based detection to handle these inference results. Specifically, in cloud, it first sends the samples to a complex model to generate a pseudo label, then it compare the inference results with these label to get a variables $acc \in \{0, 1\}$ to denote the correctness of each sample, and append them at the end of list
$W=\{acc_1,acc_2,\ldots,acc_n\}$ in cloud storage, which $n$ represent the capability of the list, it is noting that each edge will have different $W$ in cloud. Retrain managing module will used $W$ to compare the urgency degree $d$ for edge:
\begin{align}
    d &= \sum_{i=0}^{m}{\left({wa}_0-{wa}_i\right)\cdot(1/(1+e^{-i/tm})\cdot m)} \label{eq:4} \\
    wa_i &= \sum_{j=l \cdot i+1}^{(l+1) \cdot i}{acc_j} \label{eq:5}
\end{align}
In the formula, $l$ is the length of each batch $wa$, which sums up $acc$ variables, $m$ is the number of batch $wa$ in list $W$.

There we also consider temporally important differences for window $W_i$ via exponential decay weight, which is similar with the sample filtering module. Retrain manager continuously remove old samples if the number of $acc$ exceeds list capability.

\begin{algorithm}
\caption{Training Manager Algorithm}\label{alg:cap}
\begin{algorithmic}[1]
\State \textbf{Input:} samples and inference results  received from edge: $M_e=\{(x_{1e}, y_{1e}, 1), (x_{2e}, y_{2e}, 2),\ldots , (x_{re}, y_{re}, r)\}$, hyperparameters $h$ obtained in offline phase, early stop threshold $k$, complex model $f$
\State \textbf{Define:} list W's size $n$, segment's size $m$
\For{$e \in \text{edges}$}
    \For{$(x, y, i) \in M_e$}
        \State $label \gets f(x)$
        \State $acc \gets label == y$
        \State $e.\text{bank} \gets \text{append}(e.\text{bank}, (acc, i))$ \Comment{append acc and index to storage}
    \EndFor
    \If{$\text{len}(e.\text{bank}) > n$}
        \State \text{remove elements in e.bank exceeds its size n}
    \EndIf
    \State \text{divide} $e.\text{bank}$ \text{into} $m$ \text{batches}
    \State $d_e \gets \text{Score}(e.\text{bank}, m)$ \Comment{calculate degree of e via \ref{eq:4}}
\EndFor
\State $e \gets \arg\max d$ \Comment{find the edge with max $d$}
\State $epoch \gets 0$, $max\_epoch \gets 0$
\State $start\_time \gets \text{current\_time}()$, $max\_evaluation \gets 0$
\While{$epoch - max\_epoch > k$ \textbf{and} $\text{current\_time}() - start\_time > max\_time$}
    \State \text{train the model of edge} $e$ \text{with hyperparameters} $h$
    \State $evaluation \gets e.\text{eval}()$ \Comment{evaluate the performance of model of edge $e$}
    \If{$evaluation > max\_evaluation$}
        \State $max\_evaluation \gets evaluation$, $max\_epoch \gets epoch$
    \EndIf
    \State $epoch \gets epoch + 1$
\EndWhile
\end{algorithmic}
\end{algorithm}

\subsubsection{Adapt update frequency for retraining}
Ideally, if the content of the video stream changes quickly, the system should adapt to the current scene quickly, then allocate more time  on retraining model deployed on this edge and quickly send back the trained model, in order to reduces the update window size. Model profiling described above can help distinguish change degree to optimize time allocation, but it cannot control the model training time, so it is crucial to let retrain managing module dynamically adjust the training time.

Ekya shows that different training hyperparameters have a non-negligible impact on the accuracy, so we first need to determine the appropriate training hyperparameters. We found that Ekya's micro-profiler is not suitable for our system, for two main reasons. Firstly, our system requires more frequent model updates, leading to a higher proportion of time spent on online profiling methods within each update cycle. Secondly, Ekya performs training and inference simultaneously, necessitating consideration of numerous hyperparameters, whereas we require only a subset of these hyperparameters for training. We have determined that an offline profiling method is adequate for achieving desirable outcomes. This approach allows us to focus exclusively on determining retraining times for the currently selected edge model during online operations, thereby shortening the total update life cycle.

Specifically, use we use Bayesian Hyperparameter Optimization(BHO) in offline phase to efficiently explore training hyperparameters that achieve the best performance. Central to BHO are the objective, prior, and acquisition functions. The objective function evaluates the predictive performance of a model for a given hyperparameter set. The prior function encapsulates initial beliefs about the hyperparameter space, which is updated upon the acquisition of new data. Specifically, We use Gaussian Process as the prior function: $f(h_{1:k}) \sim \mathcal{GP}(m(h_{1:k}), k(h_{1:k}, h_{1:k}))$, where $m$ denotes the mean function, $k$ characterizes the covariance function modeling the relationship among input hyperparameters $h_1$ to $h_k$ and $f$ is objective function. We adopt Expected Improvement (EI) as our chosen approach for the acquisition function: $\text{EI}(h) = \mathbb{E} \left[ \max(f(h) - f_{\text{best}}, 0) \right]$ considering its high performance in our tasks.

In the offline phase, we first initially collect a set of video data covering various different scenarios, then use BHO to get a unique set of hyperparameters for each video. Then  we use the mean of these hyperparameters as the starting point, $h_0$. Following this, we randomly sample segments from each video and combine them to update the hyperparameters, Retrain Managing module stops BO when the improvement is less than a threshold. It is noting that in the process of training, we freeze the backbone and feature extraction layers and merely tune the parameters of the last layer and classification prediction layer, this approach is based on the understanding that the model's representations evolve from generic characteristics (like patterns and color gradients) to ones that are more tailored to the specific task (such as identifying objects) with the increase in layer depth, as documented in \cite{bau2017network}\cite{guo2019spottune}. 

In online phase, we use an early-stopping strategy to dynamically adjust the number of training rounds: record the validation loss in each training, if the k-th epoch has passed and the loss has not been reduced, then stop the training phase. Inspire from the Just-In-Time (JIT)\cite{mullapudi2019online}, we define a maximum allowable training time threshold and stop training immediately upon surpassing this threshold. This measure is designed to prevent the issue of long training rounds encountered in certain scenarios. This not only able to reduce the time of updating the model but also enhance the model's generalizability\cite{ji2021early}, reducing the misleading of the noise sample to the model. Algorithm 2 describes overall procedure.

\subsubsection{Send retrained parameters back to edge}
After finishing training, Retrain Managing module returns the model to the relevant edge. It's important to note that only the updated parameters, not the entire edge model, are transmitted via the network. Moreover, to speed up model updates and decrease context switching delays between the GPU and memory, the model and frozen parameters are permanently stored in the edge's GPU memory.

\section{experiments}

\subsection{Experiment Settings}
\subsubsection{Datasets}

We evaluate EdgeSync using 22 videos collected from YouTube, with a frame rate of 30 frames per second. Considering our objective of incorporating meaningful data drifts into our video, we adopt videos that have a length of at about 20 minutes, then bind them with a total video length of about 7 hours. These videos have different light intensities (day or night), clarity conditions (sunny, rainy, snowy)and varying speeds (walking, driving), thus contain a variety of data drift scenarios with varying degrees of difficulty. We randomly select two frames per second to convert these videos into images and label them with a chronological order; we distribute this data equally to each camera when multiple cameras are present in each experiment, we do not use a video twice to ensure these cameras have different scenes from each other. Note that we do not use common public datasets such as Cityscape\cite{Cordts_2016_CVPR}. This is because we want the video data to contain multiple complex scenes under the condition that the timing relationship is correct, and at the same time, we ensure that the length of a single video is not too small.

\subsubsection{Compared Approaches}
To verify the effectiveness of EdgeSync, we consider following methods:
\begin{itemize}
\item No Adaptation: Without any adaptation, we execute the pre-trained model on the edge device.

\item One-Time Adaptation\cite{rebuffi2017icarl}: We fine-tune those edge models using the first 100 seconds of samples at the beginning of each video like transfer learning, after which no further adjustments to the parameters are made. This can be seen as a baseline of not having continuous adaptation.

\item AMS\cite{khani2021real}: The overall architecture of AMS is the same as EdgeSync, which keeps repeating the training and updating of the edge model with the help of the cloud, it uses a fixed epoch to train the edge model and updating a small fraction of model parameters, in our experiments, we update the parameters of last feature layer and final classification layer for a more fair comparison.

\item Ekya\cite{bhardwaj2022ekya}: Ekya place both inference and retraining tasks on edge servers, it uses a fixed period of time as window, and using a micro-profiler to determine the resource allocation for each window. We place retraining task and the micro-profiler in the cloud to determine training hyper-parameters for current window with the size of 200s, for micro-profiler, we consider number of layers to retrain, the fraction of data between retraining windows to use for retraining, momentum in SGD and weight decay, then set fraction of training data and Early termination epoch to 20\% and 5 respectively.
\end{itemize}


\subsubsection{Models}
On edge devices, we utilize MobileNetV2 \cite{Sandler_2018_CVPR}, which demonstrates real-time inference speeds (30 frames-per-second). This efficiency holds true even on edge devices with lower computational power, such as the NVIDIA Jetson Nano and Jetson TX2. Conversely, on the cloud server, we leverage a high-performance golden model, ResNeXt101\cite{wang2019elastic}, to acquire accurate ground truth labels. Both of them are pretrained on ImageNet datasets\cite{deng2009imagenet}.

\subsubsection{Implementation Details}
We adopt the NVIDIA GeForce RTX 2080Ti GPU as our edge device. In order to enable the edge device to simulate the real scenario, we calculate the time interval between two model updates, after which we reduce the actual inference speed of the edge device in this time interval based on the FPS of the Jetson Nano. We conduct all evaluations using a single NVIDIA Tesla V100 GPU in the cloud. The models on both the edge and cloud sides are implemented in Python 3.8 and PyTorch 2.0 with CUDA 11.7. At the edge, we apply Algorithm 1 with parameters $k = 0.7$, signifying that 30\% of samples are filtered in the current window. Additionally, we set $\alpha = 1.0$ and $\beta = 1.0$. In Algorithm 2, we configure the patience parameter to 5, while the memory bank capacity $N$ and segment size $m$ are set to 90 and 10, respectively.

 

%

\begin{table}[!t]
\renewcommand{\arraystretch}{1.3}
\caption{Comparing end-to-end performance and bandwidth consumption of different methods.}
\label{table_e2e}
\centering
\begin{tabular}{cccc}
\toprule
Method & Accuracy & Data Upload & Data Download \\
\midrule
No Adaptation & 62.40\% & 0.0Kbps & 0.0Kbps \\
One-Time Adaptation & 63.96\% & 42.47Kbps & 11.13Kbps \\
AMS & 68.87\% & 254.84Kbps & 352.74Kbps \\
Ekya & 68.70\% & 254.84Kbps & 358.4Kbps \\
EdgeSync & 72.09\% & 178.39Kbps & 1.836Mbps \\
\bottomrule
\end{tabular}
\end{table}


\begin{table*}[!t]
\renewcommand{\arraystretch}{1.3}
\caption{Time spent of different methods within a cycle.}
\label{table_time}
\centering
\begin{tabular}{ccccccc}
\toprule
Method & Label time & Retraining time & Model profiling time & Network communication time & Total time \\
\midrule
EdgeSync & 9.02s & 8.77s & 1ms & 3.52s & 21.311s \\
EdgeSync* & 9.67s & 13.72s & 0.1ms & 3.52s & 26.91s \\
AMS & 43.83s & 66.25s & 0.1ms & 3.52s & 113.6s \\
Ekya & 33.2s & 58.49s & 7.84s & 3.52s & 103.05s \\
\bottomrule
\end{tabular}
\end{table*}


\begin{figure}[!t]
\centering
\includegraphics[width=2.72in]{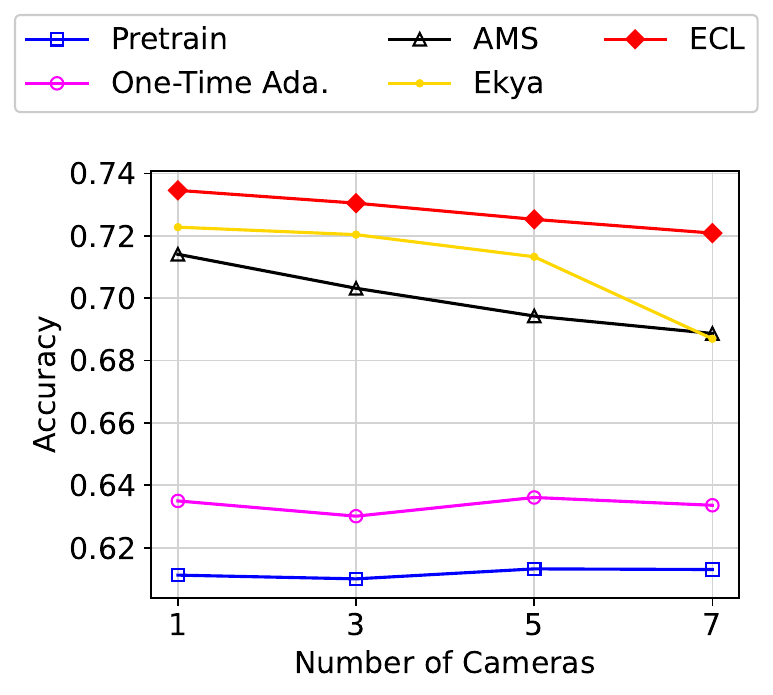}
\caption{Impact of accuracy with varying numbers of cameras.}
\label{fig_num}
\end{figure}

\subsection{Results and Discussions}
\subsubsection{Comparison to baselines}
We first compare the end-to-end accuracy of EdgeSync with the baselines which 7 number of concurrent cameras replaying videos from our dataset, Table \ref{table_e2e} shows the overall accuracy of thees five methods. To assess various methods' accuracy, we opt for a classification task, comparing edge device inference results with labels from the teacher model for video frames. Accuracy is measured as the ratio of correct predictions to total cases across six categories: people, bicycles, cars, motorcycles, buses, and trucks. From the results, it can be seen that continual training the edge model provides significant accuracy improvement, EdgeSync performs the best among others, with an 8\% increase in accuracy compared to No Adaptation method; No Adaptation has the lowest accuracy due to the lack of adaptation to a specific scene; One-Time Adaptation has slightly better accuracy than no-adjustment, but the overall accuracy is poor due to the difficulty of adapting to the future video content; the AMS continues to utilize the samples from the most recent period of time for training, so the edge model is able to maintain a certain degree of accuracy, which results in 4.9\% increase in accuracy compared to 4.9\% improvement over One-Time Adaptation; The accuracy of Ekya is comparable to that of AMS, although the performance relies on different ideas. While AMS dedicates the entirety of the cloud's time to model training and sample labeling, applying a uniform treatment to each edge model, Ekya adopts a dynamic approach. It considers the impact of current window configurations on accuracy during continuous training, strategically selecting more suitable configurations at a minor cost to the window's start time. This approach becomes particularly valuable when the cloud has to manage numerous edge tasks, as the time spent on configuration selection becomes a larger proportion of the overall time. Consequently, this finally leads to a limited enhancement of the overall accuracy of the edge.

\begin{figure}[!t]
\centering
\subfloat[Verification accuracy]{\includegraphics[width=1.55in]{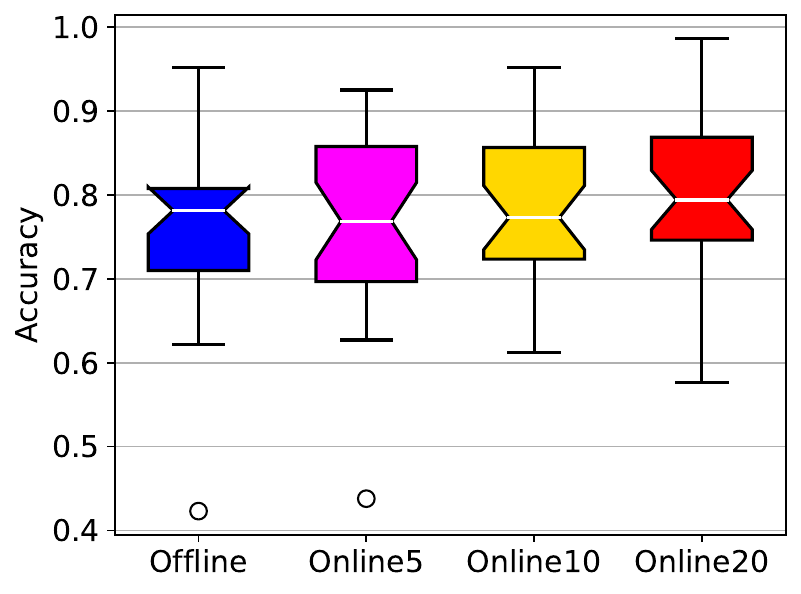}%
\label{fig_first_case}}
\hfil
\subfloat[Hyper-parameters profiling cost]{\includegraphics[width=1.9in]{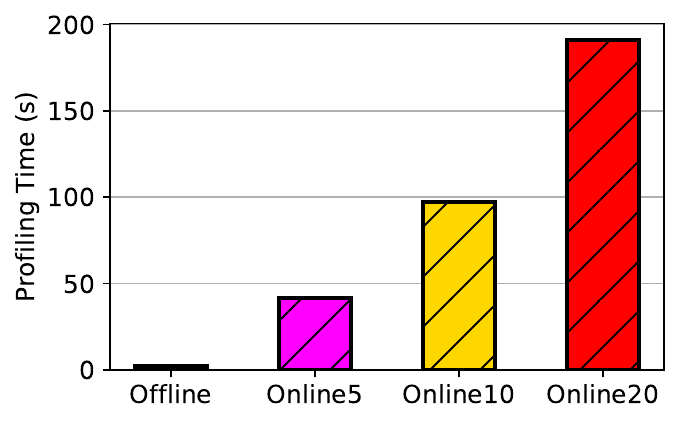}%
\label{fig_second_case}}
\caption{Impact of hyper-parameters profiling in retraining windows.}
\label{fig_ver}
\end{figure}

\subsubsection{Time Spent} Table \ref{table_time} shows how long it takes, on average, to update a model using different methods for running various components within a single update window. EdgeSync* represents the EdgeSync method without using the retraining manager module, the time cost of each retraining process consists of four parts: labeling, edge model profiling, retraining and edge-cloud communication. EdgeSync takes lowest average total time compare to other methods, The time costs for the four components in EdgeSync  are 42.32\%, 41.15\%, 0.01\%, and 16.52\%, respectively. as discussed before, reducing the time required for model updating can prevent the retrained model from becoming outdated, thus adapts to complex environments. If we remove the retraining manager module in EdgeSync, it will result in a longer single update time by 26.27\%, attributed to the rise in sample labeling and extended training duration. AMS takes 43.83s for labeling and 66.25s for retraining, Since the sample time horizon of AMS training is longer, the trained model for edge has better generalization, but when the edge video scene changes rapidly, the adaptability of AMS will decrease. Ekya employs a time-consuming heuristic that evaluates each pair of candidates to identify the pair that can enhance accuracy the most. According to the table, the profiling for every edge consumes 7.84 seconds, constituting a significant portion of the total runtime (26\% in our experimental setup), thereby imposing a substantial additional computational overhead, moreover, the model can only be updated once within a window, which is not conducive to edges demanding frequent updates.

\begin{table}[!t]
\renewcommand{\arraystretch}{1.3}
\caption{Ablation experimental performance of different modules of EdgeSync.}
\label{table_abEdgeSync}
\centering
\begin{tabular}{cccccc}
\toprule
Method & EdgeSync & AMS & EdgeSync/STF & EdgeSync/TF & EdgeSync/F \\
\midrule
Accuracy & 0.7210 & 0.6880 & 0.6840 & 0.6950 & 0.7108 \\
\bottomrule
\end{tabular}
\end{table}

\subsubsection{Impact of number of cameras} Fig \ref{fig_num} further demonstrates the effect of the number of cameras on the overall accuracy. Since both No Adaptation and One-Time Adaptation methods do not have continuous tuning during task execution, they do not need to compete for occupying the cloud resources, the accuracies of these two methods change slightly due to the differences between the video data in the cameras; in contrast, the overall accuracies of the AMS, Ekya, and EdgeSync show a decreasing trend with the increase of the number of cameras, this is because when the number of cameras increases, each lightweight model retraining frequency and the overall training time decreases, as a result, the model at the edge has low adaptability to its current video content. Compared with the other two methods, The accuracy of EdgeSync decreases more slowly than the other two methods(1.2\% less than AMS and 2.2\% less than Ekya) when the number of cameras increases, this distinction arises from the fact that, unlike Ekya and AMS, EdgeSync autonomously decides when to halt the training process. This adaptive approach increases the frequency of model updates by reducing training time while ensuring adequate training duration. Additionally, EdgeSync leverages retraining samples containing the most informative gradients, enhancing the current lightweight model's suitability for the prevailing video distributions.

\begin{figure}[!t]
\centering
\includegraphics[width=3.1in]{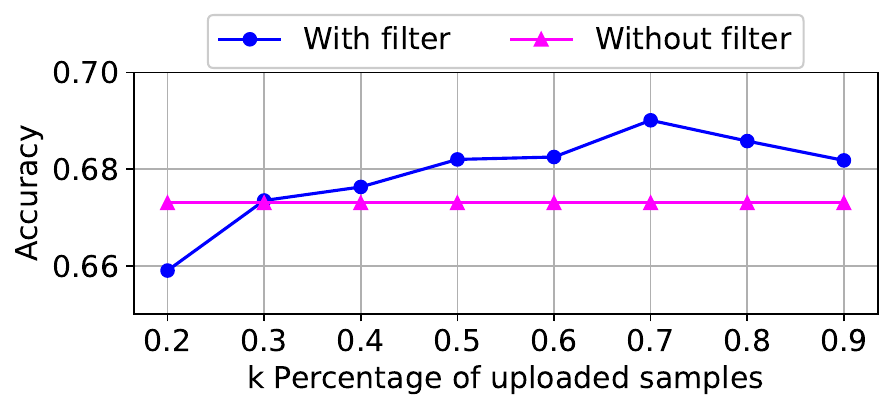}
\caption{Sensitivity analysis of filter percentage parameters.}
\label{fig_k}
\end{figure}

\subsubsection{Offline hyper-parameters profiling performance.} To demonstrate the effectiveness of the offline hyper-parameters profiling method, we compare it with the online profiling method. In this validation experiment, online selection uses the Tree-structured Parzen Estimators algorithm\cite{bergstra2011algorithms} to optimize the learning rate, momentum, and l2 penalty term, we set its number of samples parameter to 5, 10, and 20, this parameter represents the repeated times to run the search algorithm. figure \ref{fig_ver} Compares online dynamic profilings of training hyperparameters with offline profiling of fixed hyperparameter validate accuracy results and their spent time. We observe that if number of samples are too small, it performs worse than offline profiling, a long search process is required if we want to obtain better hyper-parameters, when number of samples parameter is 10, The accuracy from online profiling is very small is similar with offline method(0.1\% higher than offline profiling over 31 consecutive windows), however, each online profiling consumes an additional 100 seconds, which has a significant negative impact on the speed of model updates.

\subsubsection{Effectiveness of Each module.} In this experiment, we systematically incorporated each method individually to assess if there was an improvement in accuracy. Specifically, EdgeSync/STF denotes EdgeSync without model selection and filtering samples at the edges, it sets a fixed update time, which is the same as the window averaging time of the EdgeSync method in this experiment; EdgeSync/TF denotes EdgeSync without dynamic training time and filtering samples at the edges; EdgeSync/F represents EdgeSync without filtering samples at the edges. As shown in Table \ref{table_abecl} EdgeSync/STF demonstrates a decrease in accuracy of approximately 3.6\% compared to EdgeSync, and a slight decrease 0.4\% compared to AMS. This observation suggests that solely diminishing the model update time in AMS compromises accuracy, indicating that the model updates in AMS are prone to overfitting, leading to a significant decline in accuracy when there is a shift in the distribution of frames. EdgeSync/TF achieves around 0.5\% higher accuracy than EdgeSync/STF, which indicates the importance of profiling edge models to select the most impacted one before retraining, EdgeSync/F achieves around 1\% higher accuracy than EdgeSync/TF, demonstrating the effectiveness of dynamically deciding update time during runtime. EdgeSync/F demonstrates a decrease in accuracy of approximately 1\% compared to EdgeSync since they missing The impact degree of different samples for current model.

\subsubsection{Influence of upload percentage $k$.} figure \ref{fig_k} presents the overall accuracy and single training time across various sample filtering percentages. To mitigate potential accuracy inflation arising from increased model update frequency, we conducted experiments using a fixed model update time of 100 seconds, akin to AMS with a filter sampler. Results indicate that the baseline accuracy (unfiltered samples) is equivalent to that achieved with a 30\% sample filtering. However, when a small number of samples, such as 20\% of the data, is selected, performance suffers due to insufficient training data—a necessity for regular deep learning processes. As the volume of training data increases, model performance improves. The highest overall accuracy is observed at a filter ratio of 0.7, with a subsequent decrease beyond this point. This decline may be attributed to the retention of some interfering samples, weakening the training impact, especially considering the limited capacities of a lightweight model. This underscores that sample filtering not only reduces the training time for individual models but also fosters training consistency, leading to enhanced overall prediction accuracy.

\section{Conclusion}
In this article, we introduce an efficient framework customized for real-time video analysis, wherein a remote server is employed to perpetually train and stream model updates to the edge device. Our solution aims to minimize model update latency while enhancing the quality of continuous training. Specifically, we devise a a sample filtering module, which intelligently selects samples based on both reliability and temporal significance. Moreover, we introduce a training management module that leverages the aggregated characteristics of samples over time to identify the model most likely to yield improvements, dynamically adjusting the model training duration in response to real-time demands. Through a series of rigorous experimental analyses and ablation studies, we validate the efficacy of our approach, demonstrating notable improvements in both model performance and training efficiency.


%

\appendices
\section{Proof of the First Zonklar Equation}
Appendix one text goes here.

\section{}
Appendix two text goes here.

\section*{Acknowledgment}

The authors would like to thank...

\ifCLASSOPTIONcaptionsoff
  \newpage
\fi



%

\bibliographystyle{plain} 
\bibliography{refs} 




%

\begin{IEEEbiography}{Michael Shell}
Biography text here.
\end{IEEEbiography}

\begin{IEEEbiographynophoto}{John Doe}
Biography text here.
\end{IEEEbiographynophoto}


\begin{IEEEbiographynophoto}{Jane Doe}
Biography text here.
\end{IEEEbiographynophoto}




\end{document}